\documentclass{article}

\usepackage[main, final]{neurips_2025}

\usepackage[utf8]{inputenc} 
\usepackage[T1]{fontenc}    
\usepackage{hyperref}       
\usepackage{url}            
\usepackage{booktabs}       
\usepackage{amsfonts}       
\usepackage{nicefrac}       
\usepackage{microtype}      
\usepackage{xcolor}         
\usepackage{graphicx}       
\usepackage{amsmath}        
\usepackage{algorithm}
\usepackage{algpseudocode}
\usepackage{enumitem}
\usepackage[group-separator={,},group-minimum-digits=4]{siunitx}
\usepackage{tikz} 
\usepackage{cleveref}       
\usepackage{adjustbox}
\usepackage{needspace}
\usepackage{multirow}

\definecolor{Green}{rgb}{0,0.6,0} 

\title{BLISS: Bandit Layer Importance Sampling Strategy for Efficient Training of Graph Neural Networks}

\author{%
  \textbf{Omar Alsaqa} \\
  {\fontsize{9.1pt}{12pt}\selectfont Wilfrid Laurier University} \\
  {\fontsize{9.1pt}{12pt}\selectfont Waterloo, Ontario, Canada} \\
  \texttt{o.alsaqa@gmail.com}
  \and
  \textbf{Thi Linh Hoang}  \\
  {\fontsize{9.1pt}{12pt}\selectfont Singapore Management University} \\
  {\fontsize{9.1pt}{12pt}\selectfont Singapore} \\
  \texttt{tlhoang@smu.edu.sg}
  \and
  \textbf{Muhammed Fatih Balin} \\
  {\fontsize{9.1pt}{12pt}\selectfont Georgia Institute of Technology} \\
  {\fontsize{9.1pt}{12pt}\selectfont Atlanta, GA, USA} \\
  \texttt{balin@gatech.edu}
}

\begin{document}

\maketitle

\begin{abstract}
    Graph Neural Networks (GNNs) are powerful tools for learning from graph-structured data, but their application to large graphs is hindered by computational costs. The need to process every neighbor for each node creates memory and computational bottlenecks. To address this, we introduce BLISS, a Bandit Layer Importance Sampling Strategy. It uses multi-armed bandits to dynamically select the most informative nodes at each layer, balancing exploration and exploitation to ensure comprehensive graph coverage. Unlike existing static sampling methods, BLISS adapts to evolving node importance, leading to more informed node selection and improved performance. It demonstrates versatility by integrating with both Graph Convolutional Networks (GCNs) and Graph Attention Networks (GATs), adapting its selection policy to their specific aggregation mechanisms. Experiments show that BLISS maintains or exceeds the accuracy of full-batch training.
\end{abstract}

\section{Introduction}
Graph Neural Networks (GNNs) are powerful tools for learning from
graph-structured data, enabling applications such as personalized
recommendations \cite{ying2018graph, wang2019kgat}, drug discovery
\cite{lim2019predicting, merchant2023scaling}, image understanding
\cite{han2022vision, han2023vision}, and enhancing Large Language Models (LLMs)
\cite{yoon2023multimodal, tang2023graphgpt, chen2023graph}. Architectures like
GCNs and GATs have addressed early limitations in capturing long-range
dependencies.

However, training GNNs on large graphs remains challenging due to prohibitive
memory and computational demands, primarily because considering all neighbor
nodes for each node leads to excessive memory and computational costs. While
mini-batching, common in deep neural networks, can mitigate memory issues,
uninformative mini-batches can lead to: 1) \textbf{Sparse representations}:
Nodes may be isolated, neglecting crucial connections and resulting in poor
representations. 2) \textbf{Neighborhood explosion}: A node's receptive field
grows exponentially with layers, making recursive neighbor aggregation
computationally prohibitive even for single-node mini-batches.

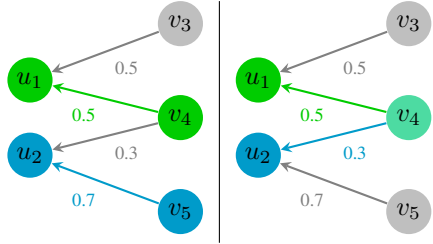
\begin{figure}[h!]
    \centering
    \begin{minipage}{0.40\textwidth}
        \centering
        \begin{tabular}{l|l}
            \begin{tikzpicture}[vertex/.style={font=\color{#1}\sffamily,circle,inner sep=0.75pt,minimum size=0.6cm}]
                \node[vertex=black,fill=green!80!black] (u1) at (0,1.5) {$u_1$};
                \node[vertex=black,fill=cyan!80!black] (u2) at (0,0.5) {$u_2$};
                \node[vertex=black,fill=gray!50!white] (v3) at (2,2.25) {$v_3$};
                \node[vertex=black,fill=green!80!black] (v4) at (2,1) {$v_4$};
                \node[vertex=black,fill=cyan!80!black] (v5) at (2,-0.25) {$v_5$};

                \draw[gray,->,>=stealth,thick,auto] (v3)-> node[font=\scriptsize] {0.5} (u1);
                \draw[green!80!black,->,>=stealth,thick,auto] (v4)-> node[font=\scriptsize] {0.5} (u1);
                \draw[gray,->,>=stealth,thick,auto] (v4)-> node[font=\scriptsize] {0.3} (u2);
                \draw[cyan!80!black,->,>=stealth,thick,auto] (v5)-> node[font=\scriptsize] {0.7} (u2);
            \end{tikzpicture}
             &
            \begin{tikzpicture}[vertex/.style={font=\color{#1}\sffamily,circle,inner sep=0.75pt,minimum size=0.6cm}]
                \node[vertex=black,fill=green!80!black] (u1) at (0,1.5) {$u_1$};
                \node[vertex=black,fill=cyan!80!black] (u2) at (0,0.5) {$u_2$};
                \node[vertex=black,fill=gray!50!white] (v3) at (2,2.25) {$v_3$};
                \node[vertex=black,fill=green!40!cyan!80!black!70] (v4) at (2,1) {$v_4$};
                \node[vertex=black,fill=gray!50!white] (v5) at (2,-0.25) {$v_5$};

                \draw[gray,->,>=stealth,thick,auto] (v3)-> node[font=\scriptsize] {0.5} (u1);
                \draw[green!80!black,->,>=stealth,thick,auto] (v4)-> node[font=\scriptsize] {0.5} (u1);
                \draw[cyan!80!black,->,>=stealth,thick,auto] (v4)-> node[font=\scriptsize] {0.3} (u2);
                \draw[gray,->,>=stealth,thick,auto] (v5)-> node[font=\scriptsize] {0.7} (u2);
            \end{tikzpicture}
        \end{tabular}
    \end{minipage}%
    \hspace{0.05\textwidth}
    \begin{minipage}{0.50\textwidth}
        \caption{The figures illustrate Node-wise vs. Layer-wise sampling. \textbf{Left:} Node-wise sampling selects nodes per target node, often causing redundancy (e.g., $v_4$ sampled for both $u_1$ and $u_2$), higher sampling rates (e.g., $v_4$, $v_5$), and missing edges (e.g., $u_2$–$v_4$). \textbf{Right:} Layer-wise sampling considers all nodes in the previous layer, preserving structure and connectivity while sampling fewer nodes.}
        \label{figure:nodevslayer}
    \end{minipage}
\end{figure}

Efficient neighbor sampling is essential to address these challenge. Techniques
include random selection, feature- or importance-based sampling, and adaptive
strategies learned during training. They fall into three categories:
\textbf{(1) Node-wise sampling}, which selects neighbors per node to reduce
cost but risks redundancy (e.g., GraphSAGE~\cite{hamilton2017inductive},
VR-GCN~\cite{chen2017stochastic}, BS-GNN~\cite{liu2020bandit}); \textbf{(2)
    Layer-wise sampling}, which samples neighbors jointly at each layer for
efficiency and broader coverage but may introduce bias (e.g.,
FastGCN~\cite{chen2018fastgcn}, LADIES~\cite{zou2019layer},
LABOR~\cite{balin2024layer}); and \textbf{(3) Sub-graph sampling}, which uses
induced subgraphs for message passing, improving efficiency but potentially
losing global context if reused across layers (e.g.,
Cluster-GCN~\cite{chiang2019cluster}, GraphSAINT~\cite{zeng2019graphsaint}).


Our key contributions are: (1) \textbf{Modeling neighbor selection as a
    layer-wise bandit problem:} Each edge represents an "arm" and the reward is
based on the neighbor's contribution to reducing the variance of the
representation estimator. (2) \textbf{Applicability to Different GNN
    Architectures:} BLISS is designed to be compatible with various GNN
architectures, including GCNs and GATs.

The remainder is organized as follows: \cref{sec:proposed_method} describes
BLISS; \cref{sec:experiments} reports results; \cref{sec:conclusion} concludes.
A detailed background and related work appear in
\cref{sec:background,sec:related_work}.

\section{Proposed Method}
\label{sec:proposed_method}
\subsection{Bandit-Based Layer Importance Sampling Strategy (BLISS)}

BLISS selects informative neighbors per node and layer via a policy-based
approach, guided by a dynamically updated sampling distribution driven by
rewards reflecting each neighbor's contribution to node representation. Using
bandit algorithms, BLISS balances exploration and exploitation, adapts to
evolving embeddings, and maintains scalability on large graphs. Traditional
node sampling often fails to manage this trade-off or adapt to changing node
importance, reducing accuracy and scalability. While \cite{liu2020bandit}
framed node-wise sampling as a bandit problem, BLISS extends it to layer-wise
sampling, leveraging inter-layer information flow and reducing redundancy (see
\cref{figure:nodevslayer}).

Initially, edge weights $w_{ij}=1$ for all $j \in \mathcal{N}_i$, with sampling
probabilities $q_{ij}$ set proportionally. BLISS proceeds top-down from the
final layer $L$, computing layer-wise sampling probabilities $p_j$ for nodes in
layer $l$. These are passed to \cref{alg:adapted_sampling}, which selects $k$
nodes. The GNN then performs a forward pass, where each node $i$ aggregates
from sampled neighbors $j_s$ to approximate its representation $\hat{\mu}_i$:

\begin{figure}[H]
    \begin{minipage}{0.38\textwidth}
        \begin{flalign} 
            \label{eq:node_prob_bliss}
             & p_j = \sqrt{\sum_{i} \left(\frac{q_{ij}}{\sum_{k \in \mathcal{N}_i} q_{ik}}\right)^2} &  &
        \end{flalign}
    \end{minipage}
    \hfill 
    \begin{minipage}{0.50\textwidth}
        \begin{flalign} 
            \label{eq:est_bliss}
             &  & h_i = \frac{1}{k} \sum_{s=1}^{k} \frac{\alpha_{{ij}s}}{q_{{ij}s}} \hat{h}_{j_s} &
        \end{flalign}
    \end{minipage}
\end{figure}
Here, $j_s \sim q_i$ denotes the $s$-th sampled neighbor of node $i$, drawn from the per-node sampling distribution $q_i$. This process updates node representations $h_j$. The informativeness of neighbors is quantified as a reward $r_{ij}$, and the
estimated rewards $\hat{r}_{ij}$ are calculated as:

\begin{figure}[H]
    \begin{minipage}{0.35\textwidth}
        \begin{flalign} 
            \label{eq:reward_bliss}
             & r_{ij} = \frac{\alpha_{ij}^2}{k \cdot q_j^2} \| h_j \|_2^2, &  &
        \end{flalign}
    \end{minipage}
    \hfill
    \begin{minipage}{0.50\textwidth}
        \begin{flalign} 
             &  & \hat{r}_{ij}^{(t)} = \frac{r_{ij}^{(t)}}{q_j^{(t)}} \quad \text{if } j \in S_i^t &
        \end{flalign}
    \end{minipage}
\end{figure}
where $S_i^t$ is the set of sampled neighbors at step $t$, $\alpha_{ij}$ is the aggregation coefficient, and $h_j$ is the node embedding. The edge weights $w_{ij}$ and sampling probabilities $q_{ij}$ are updated using the EXP3 algorithm (see \cref{alg:exp3}). The edge weights are updated as follows:

\begin{figure}[H]
    \begin{minipage}{0.38\textwidth}
        \begin{flalign} 
             & w_{ij}^{(t+1)} = w_{ij}^{(t)} \exp\left(\frac{\delta \hat{r}_{ij}^{(t)}}{|\mathcal{N}_i|}\right) &  &
        \end{flalign}
    \end{minipage}
    \hfill
    \begin{minipage}{0.50\textwidth}
        \begin{flalign}
            \label{eq:bliss_sampling_dist}
             &  & q_{ij}^{(t+1)} = (1-\eta) \frac{w_{ij}^{(t+1)}}{\sum_{j \in \mathcal{N}_i} w_{ij}^{(t+1)}} + \frac{\eta}{|\mathcal{N}_i|} &
        \end{flalign}
    \end{minipage}
\end{figure}
where $\delta$ is a scaling factor and $\eta$ is the bandit learning rate.


BLISS operates through an iterative process of four steps: (1) dynamically
selecting nodes at each layer via a bandit algorithm (e.g., EXP3) that assigns
sampling probabilities, (2) estimating node representations by aggregating from
sampled neighbors using Monte Carlo estimation, (3) performing standard GNN
message passing with these samples, and (4) calculating rewards based on
neighbor contributions to update the bandit policy and refine future sampling
distributions. For the detailed algorithm check \cref{alg:bliss}.

\subsection{Adapting to Graph Attention}
\textbf{BLISS}: We extend BLISS to attentive GNNs, following \cite{liu2020bandit}. With only a sampled neighbor set $S_i$, true normalized attention $\alpha_{ij}$ is unavailable. We compute unnormalized scores $\tilde{\alpha}_{ij}$ and define adjusted \emph{feedback} attention: $\alpha^\prime_{ij} \;=\; \sum_{j \in S_i} q_{ij}\,\frac{\tilde{\alpha}_{ij}}{\sum_{j \in S_i} \tilde{\alpha}_{ij}} $
where $q_{ij}$ is the bandit-determined sampling probability of edge $e_{ij}$. We use $\sum_{j \in S_i} q_{ij}$ as a surrogate for the normalization over the full neighborhood $N_i$, thus approximating $\alpha_{ij}$ while properly weighting sampled neighbors within the attention mechanism.

\textbf{PLADIES}:Applying LADIES to attentive GNNs (e.g., GATs) requires preserving at least one neighbor per node after sampling to respect attention’s dependence on neighbor information. The PLADIES edge-sampling procedure, adapted from \cite{balin2024layer} and detailed in \cref{alg:adapted_sampling}, first computes initial probabilities $(p_j)$, then iteratively adjusts a scaling factor $(c)$ so the sum of clipped probabilities approaches the target sample size $(k)$. Probabilities for seed nodes $V_{\text{skip}}$ are set to $\infty$, guaranteeing selection and creating ``skip connections,'' ensuring each node retains a neighbor for attention while enabling LADIES to leverage attention efficiently.

\section{Experiments}
\label{sec:experiments}

\subsection{Datasets}
We evaluate the performance of each method in the node prediction task on the
following datasets: Cora, Citeseer \cite{sen2008collective}, Pubmed
\cite{namata2012query}, Flickr, Yelp \cite{zeng2019graphsaint}, and Reddit
\cite{hamilton2017inductive}. More details of the benchmark datasets are given
in \cref{table:datasets}.

\subsection{Experiment Settings}
We compare BLISS with PLADIES, a strong baseline among existing layer-wise
sampling algorithms. The code for both BLISS and PLADIES is publicly
available.\footnote{The code implementation is available at:
    \url{https://github.com/linhthi/BLISS-GNN}}

\paragraph{Model and Training.} We use 3-layer GNNs (GraphSAGE and GATv2) with a hidden dimension of 256.
Models are trained with the ADAM optimizer with a learning rate of 0.002. For
bandit experiments, we set $\eta = 0.4$ and $\delta = \eta / 10^6$ to prevent
large updates.

\paragraph{Sampling Parameters.} Batch sizes and fanouts for each dataset are listed in
\cref{table:dataset_parameters}. For smaller datasets (Citeseer, Cora, Pubmed),
a small batch size is chosen to ensure the sampler does not process all
training nodes in a single step (training nodes: 120, 140, and 60
respectively). For larger datasets (Flickr, Yelp, Reddit), relatively small
batch sizes are used to accommodate limited computational resources (tested on
a P100 GPU with 16GB VRAM). An incremental fanout configuration ensures
sufficient local neighborhood aggregation: the first layer's fanout is set to
four times the batch size, and subsequent layers' fanouts are twice the
preceding layer's.

\paragraph{Evaluation.} For all methods and datasets, training is conducted 5 times with different
seeds, and the mean and standard deviation of the F1-score on the test set are
reported. The number of training steps for each dataset is specified in
\cref{table:dataset_parameters}. We run the experiments on GraphSAGE
\cite{hamilton2017inductive} and GATv2 \cite{brody2021attentive}.

\paragraph{Baseline Justification.} We compare BLISS against PLADIES from \cite{balin2024layer} because it
represents the state-of-the-art in layer-wise sampling, which is the specific
category BLISS belongs to. While other sampling methods like GraphSAINT
\cite{zeng2019graphsaint} (subgraph sampling) or GCN-BS \cite{liu2020bandit}
(node-wise bandit sampling) exist, direct comparison would require different
experimental setups or fall outside the scope of layer-wise sampling.
Our goal is to achieve accuracy comparable to full-batch training while
maintaining scalability, which PLADIES also aims for within the layer-wise
paradigm.

\subsection{Results}
\label{sec:results}

\definecolor{nipsPurple}{RGB}{145, 125, 185}
\begin{table}
\centering
\caption{Comparison of F1-scores (mean ± standard deviation) for BLISS and PLADIES samplers on six datasets using Graph Attention Networks (GAT) and GraphSAGE (SAGE) architectures.}
\label{table:performance_samplers}
\begin{adjustbox}{width=0.95\textwidth}
\begin{tabular}{ll|cc|cc|cc}
\toprule
\textbf{Dataset} & \textbf{Sampler} & \multicolumn{2}{c|}{\textbf{Train}} & \multicolumn{2}{c|}{\textbf{Validation}} & \multicolumn{2}{c}{\textbf{Test}} \\
 & & GAT & SAGE & GAT & SAGE & GAT & SAGE \\
\midrule
citeseer & BLISS & 0.927 ± 0.005 & 0.947 ± 0.013 & 0.712 ± 0.004 & 0.598 ± 0.028 & \textcolor{nipsPurple}{\textbf{0.706 ± 0.002}} & 0.580 ± 0.032 \\
 & PLADIES & 0.912 ± 0.007 & 0.963 ± 0.016 & 0.699 ± 0.008 & 0.616 ± 0.020 & \textcolor{nipsPurple}{\textbf{0.683 ± 0.005}} & 0.601 ± 0.017 \\
\cmidrule(lr){0-7}

cora & BLISS & 0.989 ± 0.002 & 0.983 ± 0.005 & 0.802 ± 0.005 & 0.785 ± 0.005 & \textcolor{nipsPurple}{\textbf{0.813 ± 0.004}} & 0.795 ± 0.009 \\
 & PLADIES & 0.989 ± 0.003 & 0.981 ± 0.005 & 0.800 ± 0.004 & 0.767 ± 0.011 & \textcolor{nipsPurple}{\textbf{0.809 ± 0.003}} & 0.772 ± 0.014 \\
\cmidrule(lr){0-7}

flickr & BLISS & 0.515 ± 0.003 & 0.516 ± 0.002 & 0.511 ± 0.003 & 0.503 ± 0.001 & \textcolor{nipsPurple}{\textbf{0.511 ± 0.002}} & 0.503 ± 0.002 \\
 & PLADIES & 0.511 ± 0.006 & 0.515 ± 0.001 & 0.507 ± 0.005 & 0.504 ± 0.001 & \textcolor{nipsPurple}{\textbf{0.507 ± 0.005}} & 0.505 ± 0.001 \\
\cmidrule(lr){0-7}

pubmed & BLISS & 0.907 ± 0.008 & 0.807 ± 0.063 & 0.748 ± 0.006 & 0.594 ± 0.047 & \textcolor{nipsPurple}{\textbf{0.731 ± 0.007}} & 0.597 ± 0.057 \\
 & PLADIES & 0.910 ± 0.008 & 0.760 ± 0.042 & 0.750 ± 0.014 & 0.571 ± 0.038 & \textcolor{nipsPurple}{\textbf{0.718 ± 0.013}} & 0.557 ± 0.042 \\
\cmidrule(lr){0-7}

reddit & BLISS & 0.953 ± 0.001 & 0.979 ± 0.001 & 0.949 ± 0.001 & 0.962 ± 0.000 & 0.949 ± 0.001 & \textcolor{nipsPurple}{\textbf{0.962 ± 0.000}} \\
 & PLADIES & 0.954 ± 0.002 & 0.979 ± 0.001 & 0.951 ± 0.001 & 0.962 ± 0.000 & 0.950 ± 0.001 & \textcolor{nipsPurple}{\textbf{0.962 ± 0.000}} \\
\cmidrule(lr){0-7}

yelp & BLISS & 0.540 ± 0.002 & 0.530 ± 0.005 & 0.538 ± 0.002 & 0.527 ± 0.005 & \textcolor{nipsPurple}{\textbf{0.540 ± 0.002}} & 0.529 ± 0.005 \\
 & PLADIES & 0.540 ± 0.002 & 0.503 ± 0.009 & 0.537 ± 0.002 & 0.501 ± 0.009 & \textcolor{nipsPurple}{\textbf{0.539 ± 0.002}} & 0.502 ± 0.009 \\
\bottomrule
\end{tabular}
\end{adjustbox}
\end{table}




Our experiments confirm that BLISS, a dynamic layer-wise sampling strategy,
consistently outperforms the PLADIES sampler across multiple benchmark datasets
and GNN architectures (GAT and GraphSAGE), as shown in
\cref{table:performance_samplers}. It is worth noting that the original LADIES
and PLADIES were designed specifically for GraphSAGE. A comparison of BLISS on
GAT against the original PLADIES (or LADIES) on GraphSAGE reveals a noticeable
advantage for BLISS (e.g., Citeseer: $70.6\%$ vs. $60.1\%$; Pubmed: $73.1\%$
vs. $55.7\%$).

The results demonstrate superior F1-scores for BLISS, particularly with GAT
models on Citeseer ($70.6\%$ vs. $68.3\%$) and Pubmed ($73.1\%$ vs. $71.8\%$).
This advantage stems from its bandit-driven mechanism, which better adapts to
evolving node importance, thereby reducing variance and improving
generalization. The performance gains are most pronounced on smaller datasets
(Cora, Citeseer, Pubmed) and on complex, heterogeneous graphs like Yelp, where
BLISS effectively captures nuanced class relationships ($52.9\%$ vs. $50.2\%$
with SAGE). On denser, more uniform graphs like Flickr and Reddit, the
performance difference is minimal. \cref{fig:val_f1}, \cref{fig:val_loss}
summarizes the F1-scores (mean ± standard deviation) and loss for both samplers
on GAT and GraphSAGE architectures.

These results validate our theoretical analysis: BLISS minimizes estimator
variance by dynamically prioritizing informative neighbors, unlike PLADIES'
static sampling which risks under-sampling critical nodes. The only noted
exception was overfitting on the Yelp dataset with GAT for both samplers, which
was unevaluated to maintain uniform experimental conditions.

\section{Conclusion}
\label{sec:conclusion}
In this work, we proposed Bandit-Based Layer Importance Sampling Strategy
(BLISS), a layer-wise sampling method for scalable and accurate training of
deep GNNs on large graphs. BLISS employs multi-armed bandits to dynamically
select informative nodes per layer, balancing exploration of under-sampled
regions with exploitation of valuable neighbors. This enables efficient message
passing and improved scalability. We demonstrated its applicability to diverse
GNNs, including GCNs and GATs, and presented an adaptation of PLADIES for GATs.
Experiments show BLISS matches or exceeds state-of-the-art performance while
remaining computationally efficient. Future directions include exploring
advanced bandit algorithms (e.g., CMAB) and extending BLISS to domains such as
GNN-augmented LLMs and vision tasks.



\medskip

{
    \small
    \bibliographystyle{plainnat}
    \bibliography{references}
}

\newpage

\appendix

\section{Tables}
\subsection{Notation Summary}
\begin{table}[h!]
  \centering
  \caption{Summary of key notations used throughout the paper, including their descriptions and contexts. This table serves as a reference for understanding the mathematical formulations and algorithms presented in the paper.}
  \label{table:notation_summary}
  \begin{adjustbox}{width=1.0\textwidth}
    \begin{tabular}{@{}lll@{}}
      \toprule
      \textbf{Symbol}  & \textbf{Description}                                                       & \textbf{Context} \\
      \midrule
      $\alpha_{ij}$    & Edge weight between node $i$ and $j$                                       & GNN              \\
      $\hat{h}_{j}$    & Feature vector of sampled node $j$                                         & GNN              \\
      $\mathcal{N}(i)$ & Set of neighbors of node $i$                                               & GNN              \\
      $k$              & Sample size (number of neighbors)                                          & Sampling         \\
      $q_i$            & Probability distribution of sampling node from neighbors of $i$            & Sampling         \\
      $\hat{\mu}_i$    & Estimated representation for node $i$                                      & Sampling         \\
      $q_{ij}$         & Sampling probability of neighbor node $j$ from node $i$,                   & Sampling         \\
      $p_j$            & Sampling probability of a node $j$ from all nodes of the current layer     & Sampling         \\
      $s$              & Index of the sample neighbor                                               & Sampling         \\
      $j_s$            & $s$-th sampled neighbor of node i                                          & Sampling         \\
      $w_{ij}$         & Edge weight between nodes $i$ and $j$ learned through the Bandit Algorithm & BLISS/Bandit     \\
      $r_{ij}$         & Reward of edge $e_{ij}$                                                    & BLISS/Bandit     \\
      $\eta$           & Learning rate for EXP3 algorithm                                           & BLISS/Bandit     \\
      $\delta$         & Scaling factor for EXP3 algorithm                                          & BLISS/Bandit     \\
      $S_i^t$          & Set of sampled neighbors of node i at iteration t                          & BLISS/Bandit     \\
      $c$              & Scaling factor in Poisson Sampling                                         & PLADIES          \\
      $\epsilon$       & Tolerance used in Poisson Sampling                                         & PLADIES          \\
      $n_{ref}$        & Refinement factor used in Poisson Sampling                                 & PLADIES          \\
      \bottomrule
    \end{tabular}
  \end{adjustbox}
\end{table}

\subsection{Dataset}
\begin{table}[h!]
\centering
\caption{Summary of the datasets used in the experiments, including the number of nodes, edges, features, classes, and the split ratio for training, validation, and testing. This table provides an overview of the graph properties and complexity of each dataset, highlighting the diversity in scale and structure.}
\label{table:datasets}
\begin{adjustbox}{width=0.85\textwidth}
\begin{tabular}{l|cccc|ccc}
\toprule
\textbf{Dataset} & \textbf{\# Classes} & \textbf{\# Nodes} & \textbf{\# Edges} & \textbf{\# Features} & \textbf{\# Train} & \textbf{\# Validation} & \textbf{\# Test} \\
\midrule
Cora & \num7 & \num{2708} & \num{10556} & \num{1433} & \num{140} & \num{500} & \num{1000} \\
Citeseer & \num6 & \num{3327} & \num{9228} & \num{3703} & \num{120} & \num{500} & \num{1000} \\ 
Pubmed & \num3 & \num{19717} & \num{88651} & \num{500} & \num60 & \num{500} & \num{1000} \\ 
\cmidrule(lr){0-7}
Flickr & \num7 & \num{89250} & \num{899756} & \num{500} & \num{44625} & \num{22312} & \num{22313} \\ 
Reddit & \num41 & \num{232965} & \num{11606919} & \num{602} & \num{153431} & \num{23831} & \num{55703} \\ 
Yelp & \num{100} & \num{716847} & \num{13954819} & \num{300} & \num{537635} & \num{107527} & \num{71685} \\
\bottomrule
\end{tabular}
\end{adjustbox}
\end{table}
\begin{table}[h!]
\centering
\caption{Experiment settings for each dataset, including batch size, fanout configuration, and the number of training steps.}
\label{table:dataset_parameters}
\begin{tabular}{lccc}
\toprule
\textbf{Dataset} & \textbf{Batch Size} & \textbf{Fanouts} & \textbf{Steps} \\
\midrule
Citeseer & 32 & [512, 256, 128] & 1000 \\
Cora & 32 & [512, 256, 128] & 1000 \\
Flickr & 256 & [4096, 2048, 1024] & 1000 \\
Pubmed & 32 & [512, 256, 128] & 1000 \\
Reddit & 256 & [4096, 2048, 1024] & 3000 \\
Yelp & 256 & [4096, 2048, 1024] & 10000 \\
\bottomrule
\end{tabular}
\end{table}


\section{Background}
\label{sec:background}

We denote a directed graph $\mathcal{G} = (\mathcal{V}, \mathcal{E})$
consisting of a set of nodes $\mathcal{V} = \{v_i\}_{i=1:N}$ and a set of edges
$\mathcal{E}=\{e_{ij} | j \in \mathcal{N}_i\}_{i=1:N} \subseteq \mathcal{V}
    \times \mathcal{V}$, where $N$ is the number of nodes, $\mathcal{N}_i$ denotes
the set of neighbors of node $v_i$, and $L$ is the number of layers.

\textbf{Graph Neural Networks.} GNNs operate on the principle of neural message passing \cite{gilmer2017neural}, where nodes iteratively aggregate information from their local neighborhoods. In a typical GNN, the embedding of node $v_i$ at layer $l+1$ is computed from layer $l$ as follows:
\begin{equation}
    \label{eq:GNN}
    h_i^{(l+1)}=\sigma\left(\sum_{j\in \mathcal{N}(i)} {\alpha^{(l)}_{ij} W^{(l)}h^{(l)}_j}\right)
\end{equation}
where $W^{(l)}$ is a learnable weight matrix, $h_j^{(l)}$ is the node feature vector at layer $l$, and $\sigma$ is a non-linear activation function. The term $\alpha_{ij}$ represents the aggregation coefficient, which varies depending on the GNN architecture (e.g., static in GCNs \cite{kipf2016semi} or dynamic in GATs \cite{velickovic2017graph}).

\textbf{Layer-wise Sampling.}\\
Following \cite{huang2018adaptive}, \cref{eq:GNN} can be written in expectation form:
\begin{equation}
    \label{eq:GNN-ep}
    h_i^{(l+1)} = \sigma_{W^{(l)}} \left(N\left(i\right)\mathbb{E}_{p_{ij}}\left[ h_j^{(l)}\right]  \right)
\end{equation}
where $p_{ij} = p(v_j|v_i)$ is the probability of sampling $v_j$ given $v_i$, and $\mathcal{N}(i) = \sum_j \alpha_{ij}$. To make the computation of \cref{eq:GNN-ep} tractable, the expectation $\mu_p (i) = \mathbb{E}_{p_{ij}}[h_j^{(l)}]$ can be approximated via Monte-Carlo sampling:
\begin{equation}
    \label{eq:node-wise}
    \hat{\mu}_p(i) = \frac{1}{n}\sum_{j=1}^n \hat{h}_{{j}}^{(l)}, {\hat{v}_j} \sim p_{ij}
\end{equation}
\cref{eq:node-wise} defines \textit{node-wise sampling}, where neighbors are recursively sampled for each node. While this reduces immediate computational load, the receptive field still grows exponentially with network depth $d$, leading to $O(n^d)$ dependencies in the input layer for deep networks. An alternative approach is to apply importance sampling to \cref{eq:GNN-ep}, which forms the basis for \textit{layer-wise sampling} methods:
\begin{equation}
    \label{eq:GNN-ep-ip}
    h_i^{(l+1)} = \sigma_{W^{(l)}} \left(N\left(i\right)\mathbb{E}_{q_j}\left[\frac{p_{ij}}{q_j}h_j^{(l)}\right] \right)
\end{equation}
where $q_j = q(v_j|v_1, ..., v_n)$ is the probability of sampling node $v_j$ from the entire layer. We estimate the expectation $\mu_q(i)$ via Monte-Carlo sampling:
\begin{equation}
    \label{eq:layer-wise}
    \hat{\mu}_q(i) = \frac{1}{n}\sum_{j=1}^n \frac{p_{ij}}{q_j} \hat{h}_j^{(l)}, \hat{v}_j \sim q_j
\end{equation}
The embedding then becomes $h_i^{(l+1)} = \sigma_{W_{(l)}}\left(\mathcal{N}(i)\hat{\mu}_q(i)\right)$. Without loss of generality, following the setting from \cite{liu2020bandit}, we assume $p_{ij} = \alpha_{ij}$ and normalize the probabilities such that $\mathcal{N}(i) = 1$. We denote $\mu_q(i)$ as $\mu_i$ for simplicity and ignore non-linearities. The goal of a layer-wise sampler is to approximate:
\begin{equation}
    \label{eq:layer-wise-norm}
    \begin{aligned}
        h_i^{(l+1)} = \hat{\mu}_i  =  \frac{1}{n}\sum_{j=1}^n \frac{\alpha_{ij}}{q_j} \hat{h}_j^{(l)}, \hat{v}_j \sim q_j
    \end{aligned}
\end{equation}
An effective estimator should minimize variance. The variance of the estimator in \cref{eq:layer-wise-norm} is:
\begin{equation}
    \label{eq:variance}
    \begin{aligned}
        \mathbb{V}(\hat{\mu}_i) & = \mathbb{E}\left[\left(\hat{\mu}_i - \mathbb{E}\left[\hat{\mu}_i\right]\right)^2\right]
        = \mathbb{E}\left[\|\hat{\mu}_i -\mu_i\|^2\right]                                                                                                                       \\
                                & = \mathbb{E}\left[\left\|\frac{\alpha_{ij}}{q_{j}}h_{j}^{(l)} - \sum_{\substack{s \in \mathcal{N}_i}} \alpha_{sj}h_{j}^{(l)}\right\|^2\right]
    \end{aligned}
\end{equation}
We seek $q_j^{\star} \ge 0$ that minimizes $\mathbb{V}(\hat{\mu}_i)$. The optimal sampling distribution is:
\begin{equation}\label{eq:optsample}
    q_{j}^{\star} = \sqrt{\sum_i \Bigg(\frac{\alpha_{ij}\left\|h_{j}^{(l)}\right\|_2}{\sum_{s\in\mathcal{N}_i}\alpha_{sj}\left\|h_{j}^{(l)}\right\|_2} \Bigg)^2} 
\end{equation}


\section{Related Work}
\label{sec:related_work}

To address these challenges, efficient neighbor sampling techniques are
crucial. These methods typically involve randomly selecting a fixed number of
neighbors, sampling based on node features or importance scores, or employing
adaptive strategies that learn optimal sampling during training. They can be
broadly categorized into three groups: \textbf{1) Node-wise sampling} samples a
subset of neighbors for each node. While this reduces immediate computations
and memory usage, the recursive nature can introduce redundancy. Examples
include GraphSAGE \cite{hamilton2017inductive}, VR-GCN
\cite{chen2017stochastic}, and BS-GNN \cite{liu2020bandit}. \textbf{2)
    Layer-wise sampling} jointly selects neighbors for all nodes at each layer,
potentially offering better efficiency and capturing broader relationships than
purely node-wise methods. However, it can introduce biases if certain graph
parts are consistently under-sampled. Examples include FastGCN
\cite{chen2018fastgcn}, LADIES \cite{zou2019layer}, and LABOR
\cite{balin2024layer}. See \cref{figure:nodevslayer} for a visual example.
\textbf{3) Sub-graph sampling} focuses on smaller, self-contained induced
subgraphs for message passing. While efficient, using the same subgraph across
all layers risks losing global context. Examples include Cluster-GCN
\cite{chiang2019cluster} and GraphSAINT \cite{zeng2019graphsaint}.

\subsection{GNN Architectures} \label{subsec:GNN-Arch}
\textbf{Graph Convolutional Networks (GCNs)} leverage a simplified convolution operation on the graph, aggregating information from a node's neighbors, and produces the normalized sum of them as in \cref{eq:GNN}
where $\sigma$ is an activation function (ReLU for GCNs), $\mathcal{N}(i)$ is the set of its one-hop neighbors, $\alpha^{(l)}_{ij}=\frac{1}{c_{ij}}$, $c_{ij}=\sqrt{|\mathcal{N}(i)|}\sqrt{|\mathcal{N}(j)|}$, $W^{(l)}$ is the weight matrix for the $l$-th layer, $h_j^{(l)}$ denotes the node feature matrix at layer $l$. For \textbf{GraphSAGE} is $c_{ij}=|\mathcal{N}(i)|$. This equally weights the contributions from all neighbors.
\\
\textbf{Graph Attention Networks (GATs)} address the equal contribution by introducing an attention mechanism that assigns learnable weights ($\alpha$) to each neighbor based on their features, allowing the model to focus on the most relevant information:
\begin{equation}
    \label{eq:gat-linear}
    e_{ij}^{(l)} = \text{LeakyReLU}(\vec a^{(l)^\top}(W^{(l)}h_i^{(l)}||W^{(l)}h_j^{(l)}))
\end{equation}
this computes a pair-wise un-normalized attention score between two neighbors. It first concatenates the linear transformation of $l$-th layer embeddings of the two nodes, where $||$ denotes concatenation, then takes a dot product of it and a learnable weight vector $\vec a^{(l)}$, and applies a LeakyReLU in the end.
\begin{equation}
    \alpha_{ij}^{(l)} = \frac{\exp(e_{ij}^{(l)})}{\sum_{k\in \mathcal{N}(i)}^{}\exp(e_{ik}^{(l)})}
\end{equation}

This difference allows GATs to capture more nuanced relationships within the
graph than GCNs. However, \cite{brody2021attentive} argues that the original
GAT uses a static attention mechanism due to the specific order of operations
in \cref{eq:gat-linear}. While the weights depend on both nodes, this structure
can limit the expressiveness of the attention calculation. GATv2 introduces a
dynamic attention mechanism by modifying this order. This change allows the
attention weights to depend on the features of both the sending node (neighbor)
and the receiving node in a potentially more expressive way. The key difference
lies in the order of operations. \cref{eq:gat-linear} will be changed to:
\begin{equation}
    e_{ij}^{(l)} = \vec a^{(l)^\top} \text{LeakyReLU}(W^{(l)}[h_i^{(l)}||h_j^{(l)}])
\end{equation}

\subsection{Sampling Technique}

\begin{algorithm}
    \caption{Sampling Procedure of LADIES}
    \label{alg:ladies}
    \begin{algorithmic}[1]
        \Require Normalized edge weights $\alpha_{ij}$; Batch Size $b$, Sample Number $k$;
        \State Randomly sample a batch of $b$ output nodes.
        \For{$l \leftarrow L$ to $1$}
        \State Calculate sampling probability for each node using $p_{j}^{l}$ in \cref{eq:prob-ladies}
        \State Sample $k$ nodes in $l$-th layer using $p_{j}^{l}$.
        \State Normalize the edge weights of the sampled nodes in the layer by \cref{eq:norm-prob-ladies}.
        \EndFor
        \State \textbf{return} Modified edge weights $\tilde{\alpha}_{ij}$ and Sampled Nodes;
    \end{algorithmic}
\end{algorithm}

This section reviews existing sampling techniques and discusses their
limitations. \\ \textbf{Layer-Dependent Importance Sampling (LADIES)
    \cite{zou2019layer}:} LADIES leverages layer-wise importance scores based on
node features and graph structure to guide node selection. LADIES begins by
selecting a subset of nodes in the upper layer. For each selected node, it
constructs a bipartite subgraph of its immediate neighbors. It then calculates
importance scores for these neighbors and samples a fixed number based on these
scores. This process is repeated recursively for each layer. However, LADIES
relies on pre-computed importance scores, which can be computationally
expensive and may not adapt well to dynamic edge-weight changes. Additionally,
LADIES employs sampling with replacement, which can be suboptimal as it may
select the same node multiple times. While LADIES suggests using $\alpha_{ij}$
values similar to GraphSAGE ($c_{ij}=|\mathcal{N}(i)|$), their implementation
utilizes \cref{eq:norm-prob-ladies} for normalization. Instead of directly
feeding $\alpha_{ij}$ to the model, it is first used to calculate an importance
score:
\begin{equation}
    \label{eq:prob-ladies}
    p_j^{(l)} = \frac{\pi_j^{(l)}}{\sum_{j} \pi_j^{(l)}}\ \text{, where}\
    \pi_j^{(l)} = \sum_{j \in \mathcal{N}_i^{(l)}} \alpha^2_{ij}
\end{equation}
The most important nodes are then sampled using $p^{(l)}_j$. Before passing these nodes to the model, the original $\alpha_{ij}$ is re-weighted by $p_{j}^{(l)}$ and normalized by dividing it over $c_{ij}$ of the selected (union of sampled nodes and seed nodes) nodes. These new $\tilde{\alpha}_{ij}^{(l)}$ values are passed to the model at each layer for the selected points.
\begin{equation}
    \label{eq:norm-prob-ladies}
    \tilde{\alpha}_{ij}^{(l)} = \frac{\alpha_{ij} / p_j^{(l)}}{\sum_{j}\left( \alpha_{ij} / p_j^{(l)} \right)} \text{,}\
    \tilde{\alpha}_{ij}^{(l)} = \frac{\alpha_{ij} / p_j^{(l)}}{c_{ij}}
\end{equation}

SKETCH \cite{chen2022calibrate} proposed a fix for the sampling equation and
the normalization of the edge weights. Instead of \cref{eq:prob-ladies}, they
suggested:
\begin{equation}
    \label{eq:prob-sketch}
    \pi_j^{(l)} = \sqrt{\sum_{j \in \mathcal{N}_i^{(l)}} \alpha^2_{ij}}
\end{equation}
SKETCH uses $\alpha_{ij}$ based on the GCN model ($c_{ij}=\sqrt{|\mathcal{N}(i)|}\sqrt{|\mathcal{N}(j)|}$).
They also suggested an alternative normalization for the edge weight instead of \cref{eq:norm-prob-ladies}:

\begin{equation}
    \label{eq:norm-prob-sketch}
    \tilde{\alpha}_{ij}^{(l)} = \frac{\alpha_{ij} / p_j^{(l)}}{ns_{j}^{(l)}}
\end{equation}
where $ns_j^{(l)}$ is the number of sampled nodes for node $i$ at layer $l$.

\noindent\textbf{Layer-Neighbor Sampling (LABOR) \cite{balin2024layer}:} The LABOR sampler combines layer-based and node-based sampling. It introduces a per-node hyperparameter to estimate the expected number of sampled neighbors, enabling correlated sampling decisions among vertices. This hyperparameter and the sampling probabilities are optimized to sample the fewest vertices in an unbiased manner.

The paper also introduced PLADIES (Poisson LADIES), which employs Poisson
sampling to achieve unbiased estimation with reduced variance. PLADIES assigns
each node $j$ in the neighborhood of source nodes $S$ (denoted $N(S)$) a
sampling probability $p_j \in [0,1]$ such that $\sum_{j \in N(S)}p_j = k$,
where $k$ is the desired sample size. A node $j$ is then sampled if a random
number $\phi_j \sim U(0,1)$ satisfies $\phi_j \leq p_j$. PLADIES achieves this
unbiased estimation in linear time, in contrast to the quadratic complexity of
some debiasing methods \cite{chen2022calibrate}. Notably, its variance
converges to 0 if all $p_j = 1$, highlighting its effectiveness.

\noindent\textbf{Bandit Samplers \cite{liu2020bandit}:} Bandit Samplers frame the optimization of sampling variance as an adversarial bandit problem, where rewards depend on evolving node embeddings and model weights. While node-wise bandit sampling is effective, selecting neighbors individually can lead to redundancy and may not capture long-range dependencies efficiently. This highlights the importance of extending to layer-wise sampling.

Their method employs a multi-armed bandit framework to learn a sampling
distribution $q_i^t$ for each node $v_i$ at each training step $t$. The
algorithm initializes a uniform sampling distribution. During training, it
samples $k$ neighbors for each node based on $q_i^t$, computes rewards based on
GNN performance, and updates the distribution using an algorithm like EXP3.
This process prioritizes informative neighbors to improve training efficiency.
Our work builds upon this foundation by applying bandit principles to the
layer-wise sampling paradigm.


\section{Complexity, Variance and Runtime}
\subsection{BLISS complexity and variance analysis}
\begin{table}[h!]
    \centering
    \caption{Summary of memory complexity, time complexity, and variance for Full-Batch, GraphSAGE, LADIES, and BLISS methods. This table provides a theoretical comparison of the computational and statistical properties of each method, emphasizing BLISS's ability to minimize variance while maintaining scalability.}
    \label{tab:complexity}
    \begin{adjustbox}{width=1.0\textwidth}
        \begin{tabular}{@{}llll@{}}
            \toprule
            \textbf{Methods}                             & \textbf{Memory Complexity}                        & \textbf{Time Complexity}                            & \textbf{Variance}                                       \\
            \midrule
            Full-Batch                                   & $O\left(L|V|K + LK^2\right)$                      & $O\left(L||A||K + L|V|K^2\right)$                   & 0                                                       \\
            GraphSage                                    & $O\left(bKs^{L-1}_{node} + LK^2\right)$           & $O\left(bKs^{L}_{node} + bK^2s^{L-1}_{node}\right)$ & $O\left(D\phi\|P\|_F^2 / (|V|s_{node})\right)$          \\
            LADIES                                       & $O\left(LK s_{layer} + LK^2\right)$               & $O\left(LK s^2_{layer} + LK^2 s_{layer}\right)$     & $O\left(\phi\|P\|_F^2 \bar{V}(b)/(|V|s_{layer})\right)$ \\

            BLISS                                        & $O\left(L|E| + LK s_{\text{layer}} + LK^2\right)$ & $O(L|E| + LK
            s_{\text{layer}}^2 + LK^2 s_{\text{layer}})$ & $O\left(\phi\|P\|_F^2 \bar{V}(b)
            / |V| s_{\text{layer}} (1-\eta)\right)$                                                                                                                                                                          \\ \bottomrule
        \end{tabular}
    \end{adjustbox}
\end{table}

\subsubsection{Memory Complexity}
\begin{itemize}
    \item  $O(L|E|)$: Stores bandit weights $w_{ij}$ for all edges across $L$ layers.
    \item  $O(LK s_{\text{layer}})$: Stores embeddings for $s_{\text{layer}}$ sampled nodes per layer ($K$-dimensional).
    \item  $O(LK^2)$: Stores $L$ weight matrices $\mathbf{W}^{(l)} \in \mathbb{R}^{K \times K}$.
\end{itemize}

\subsubsection{Time Complexity}
\begin{itemize}
    \item  $O(L|E|)$: Bandit weight updates (EXP3) over all edges in $L$ layers.
    \item  $O(LK s_{\text{layer}}^2)$: Importance score computation for $s_{\text{layer}}$ nodes per layer.
    \item  $O(LK^2 s_{\text{layer}})$: Message passing and aggregation for $s_{\text{layer}}$ nodes.
\end{itemize}

\subsubsection{Variance}
\begin{itemize}
    \item  Key Difference from LADIES: The $(1-\eta)^{-1}$ term accounts for exploration
          in bandit sampling.
    \item  Derivation: Minimizing \cref{eq:variance} with bandit-optimized $q_j$
          (\cref{eq:bliss_sampling_dist}) introduces the $\eta$-dependent denominator.
\end{itemize}

\subsection{PLADIES complexity and variance analysis}
PLADIES (Poisson LADIES) shares identical complexity terms with LADIES. The
differences between them is that PLADIES uses Poisson sampling (variable-size,
unbiased) instead of fixed-size sampling, and PLADIES reduces empirical
variance but retains the same asymptotic bound.

In the \cref{tab:complexity}, PLADIES is grouped under LADIES since their
theoretical complexities are identical. BLISS explicitly diverges due to bandit
overhead and adaptive exploration.

\subsection{Training time}
The reported time in \cref{table:time_samplers} measures per-iteration training
time - the wall-clock time taken to execute one training step. The code for
BLISS is not optimized (using naive for loops in the current implementation)
and the comparison might not be reasonable, but for the sake of the having a
clearer image about the performance. The time is also averaged over 5 runs per
experiment. \begin{table}[h!]
    \centering
    \caption{Average training time per iteration (in seconds) for BLISS and PLADIES samplers across six datasets. The table highlights the computational efficiency of both samplers, with BLISS incurring slightly higher overhead due to its dynamic bandit-based sampling mechanism, and the naive loop implementation.}
    \label{table:time_samplers}
    \begin{adjustbox}{width=1.0\textwidth}
        \begin{tabular}{ll|cc||ll|cc}
            \toprule
            \textbf{Dataset}          & \textbf{Sampler} & \multicolumn{2}{c||}{\textbf{Time}} & \textbf{Dataset} & \textbf{Sampler}        & \multicolumn{2}{c}{\textbf{Time}}                                 \\
            \cmidrule(lr){3-4} \cmidrule(lr){7-8}
                                      &                  & \textbf{GAT}                        & \textbf{SAGE}    &                         &                                   & \textbf{GAT}  & \textbf{SAGE} \\
            \midrule
            \multirow{2}{*}{Citeseer} & BLISS            & 0.065 ± 0.001                       & 0.059 ± 0.002    & \multirow{2}{*}{Pubmed} & BLISS                             & 0.722 ± 0.008 & 0.690 ± 0.007 \\
                                      & PLADIES          & 0.055 ± 0.001                       & 0.051 ± 0.002    &                         & PLADIES                           & 0.667 ± 0.008 & 0.627 ± 0.007 \\
            \cmidrule(lr){1-8}
            \multirow{2}{*}{Cora}     & BLISS            & 0.066 ± 0.001                       & 0.058 ± 0.001    & \multirow{2}{*}{Reddit} & BLISS                             & 0.207 ± 0.003 & 0.165 ± 0.007 \\
                                      & PLADIES          & 0.054 ± 0.001                       & 0.049 ± 0.001    &                         & PLADIES                           & 0.156 ± 0.002 & 0.110 ± 0.003 \\
            \cmidrule(lr){1-8}
            \multirow{2}{*}{Flickr}   & BLISS            & 0.086 ± 0.002                       & 0.080 ± 0.002    & \multirow{2}{*}{Yelp}   & BLISS                             & 0.129 ± 0.003 & 0.122 ± 0.004 \\
                                      & PLADIES          & 0.073 ± 0.002                       & 0.063 ± 0.002    &                         & PLADIES                           & 0.110 ± 0.002 & 0.102 ± 0.002 \\
            \bottomrule
        \end{tabular}
    \end{adjustbox}
\end{table}


\section{Algorithms}
\subsection{BLISS}
\begin{algorithm}
    \caption{BLISS Algorithm}
    \label{alg:bliss}
    \begin{algorithmic}[1]
        \Require Graph $G$, Sample size $k$, Bandit learning rate $\eta$, Steps $T$, Number of layers $L$
        \State Initialize $w_{ij} = 1$ if $j \in \mathcal{N}_i$ else $0$
        \For{$t = 1$ to $T$}
        \For{$l = L$ to $1$} \Comment{Top-down layers}
        \State Calculate sampling distribution $q_{ij}$, using \cref{eq:bliss_sampling_dist}
        \State Calculate node sampling probability $p_j$, using \cref{eq:node_prob_bliss}
        \State Pass the $p_j$ to \cref{alg:adapted_sampling}
        \State Sample $k$ nodes for the current layer based on $p_j$.
        \EndFor
        \State Run forward pass of GNN
        \State Get the updated node embeddings $h_j$ from \cref{eq:est_bliss} and rewards $r_{ij}$ using \cref{eq:reward_bliss}
        \State Update  wights $w_{ij}$ using EXP3 in \cref{alg:exp3}
        \EndFor
    \end{algorithmic}
\end{algorithm}

\subsection{Iterative Thinning Poisson Sampler}
\begin{algorithm}
    \caption{Iterative Thinning Poisson Sampler}
    \label{alg:adjust_scale_factor}
    \begin{algorithmic}[1]
        \Require Node probabilities $p_j$, sample size $k$, tolerance  $\epsilon$, refinement factor $n_{ref}$
        \State Initialize scaling factor $c = 1.0$
        \For{$i = 1$ to $n_{ref}$}
        \State Adjust probabilities: $S = \sum \min(p_j \cdot c, 1)$
        \If{$\min(S, k) / \max(S, k) \geq \epsilon$}
        \State \textbf{break}
        \EndIf
        \State Update scaling factor: $c = c \cdot k / S$
        \EndFor
        \State \Return $c$
    \end{algorithmic}
\end{algorithm}

\needspace{10cm}
\subsection{PLADIES}
\begin{algorithm*}
    \caption{Poisson Sampling with Skip Connections}
    \label{alg:adapted_sampling}
    \begin{algorithmic}[1]
        \Require Input subgraph $G_{\text{sub}}$, seed nodes $V_s$, edge probabilities $\alpha_{ij}$, sample size $k$, tolerance $\epsilon$, refinement factor $n_{ref}$
        \State Compute node probabilities $p_j$ based on edge weights $w_{ij}$ from \cref{eq:prob-ladies}.
        \If{$ |V_{\text{sub}}| \leq k $}
        \State \Return $ p_j \gets 1 \ \forall j \in V_{\text{sub}} $ \Comment{Include all nodes}
        \EndIf
        \State $ c\gets\textsc{IterativeThinningPoissonSampler}(p_j, k, \epsilon, n_{\text{ref}}) $ \Comment{From \cref{alg:adjust_scale_factor}}
        \State $ V_{\text{skip}} \gets \{ j \mid j \in V_s \cap V_{\text{sub}} \} $ \Comment{Identify seed nodes in subgraph}
        \For{$ j \in V_{\text{sub}} $}
        \If{$ j \in V_{\text{skip}} $}
        \State $ p_j \gets \infty $ \Comment{Ensure seed nodes are always selected}
        \Else
        \State $ p_j \gets \min(p_j \cdot c, 1) $ \Comment{Clip and scale probabilities}
        \EndIf
        \EndFor
        \State \Return $ p_j \ \forall j \in V_{\text{sub}} $
    \end{algorithmic}
\end{algorithm*}

\needspace{10cm}
\subsection{EXP3}
\begin{algorithm*}
    \caption{EXP3}
    \label{alg:exp3}
    \begin{algorithmic}[1]
        \Require Neighbor size $n$, Sample size $k$, Bandit learning rate $\eta$, Number of layers $L$
        \For{$l = L$ to $1$}
        \State Calculate $\alpha_{ij} = \sum_{j \in S_i} q_{ij} \cdot \frac{\Tilde{\alpha}_{ij}}{\sum_{j \in S_i} \Tilde{\alpha}_{ij}}$
        \State Calculate estimated rewards $\hat{r}_{ij} = \frac{r_{ij}}{p_i}$
        \State Update weights $w_{ij} = w_{ij} \exp(\frac{\delta r_{ij}}{n_i p_i})$
        \EndFor
    \end{algorithmic}
\end{algorithm*}

\section{Plots}

The advantages of BLISS are particularly pronounced in smaller datasets
(Citeseer, Cora, Pubmed) and highly heterogeneous graphs like Yelp (100
classes) with SAGE. For Yelp, BLISS achieves a test F1-score $52.9\%$ (SAGE),
while PLADIES lags at $50.2\%$. The bandit mechanism likely captures nuanced
class relationships more effectively in such complex settings. In contrast,
Flickr and Reddit exhibit minimal differences between the samplers, possibly
due to their dense connectivity and uniform class distributions, which reduce
the impact of adaptive sampling.

GAT models generally benefit more from BLISS than SAGE. For example, on Cora,
BLISS achieves a test F1-score of $81.3\%$ (GAT) compared to $80.9\%$ for
PLADIES, while SAGE shows narrower margins $79.5\%$ vs. $77.2\%$. This aligns
with our hypothesis that attention mechanisms, which dynamically weigh neighbor
contributions, synergize well with BLISS’s reward-driven sampling. SAGE’s
uniform aggregation is less sensitive to neighbor selection, though BLISS still
improves its performance.

Despite larger fanouts and batch sizes for Flickr, Reddit, and Yelp
(\cref{table:dataset_parameters}), BLISS maintains computational efficiency.
Reddit’s test F1-scores $94.9\%$ for BLISS vs. $95.0\%$ for PLADIES, highlight
that both samplers scale effectively to massive graphs, though BLISS’s adaptive
policy incurs negligible overhead. The higher step counts for Reddit (3,000)
and Yelp (10,000) reflect their size but do not compromise BLISS’s stability,
as evidenced by low standard deviations.

The Yelp dataset with GAT presented a challenge for both samplers, showing
overfitting (\cref{fig:val_loss}). While early stopping or hyperparameter
adjustments could potentially alleviate this, they were not added here to
preserve uniform experimental conditions across all datasets.

\begin{figure}[h!]
    \centering
    \makebox[\textwidth]{\includegraphics[width=0.85\textwidth]{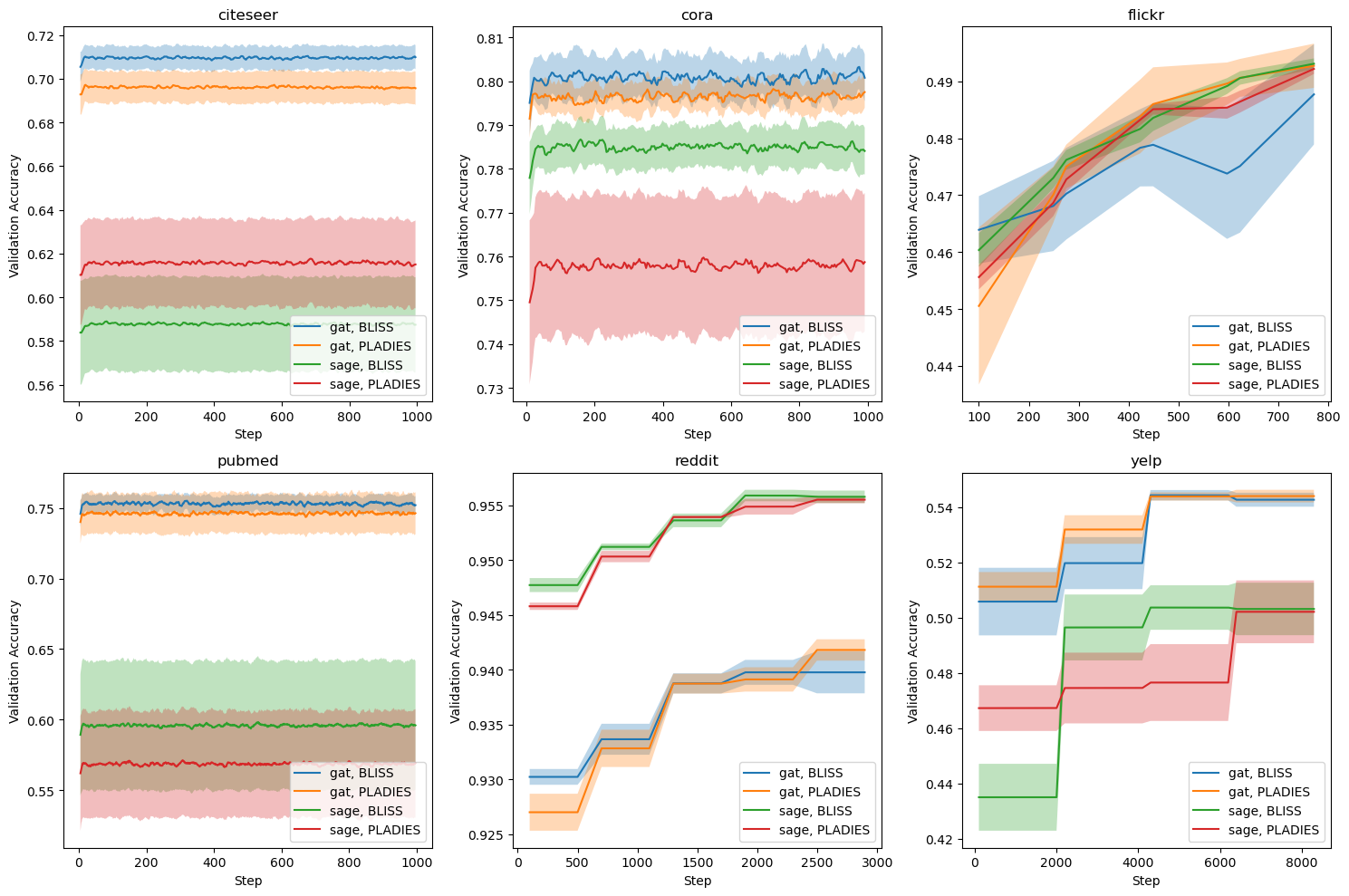}}
    \caption{Validation Accuracy across six datasets (Citeseer, Cora, Pubmed, Flickr, Yelp, and Reddit) for BLISS and PLADIES samplers using Graph Attention Networks (GAT) and GraphSAGE (SAGE) architectures. The figure highlights the performance trends during training averaged over 5 runs. The shaded regions represent the standard deviation across runs.}
    \label{fig:val_f1}
\end{figure}

\begin{figure}[h!]
    \centering
    \makebox[\textwidth]{\includegraphics[width=0.85\textwidth]{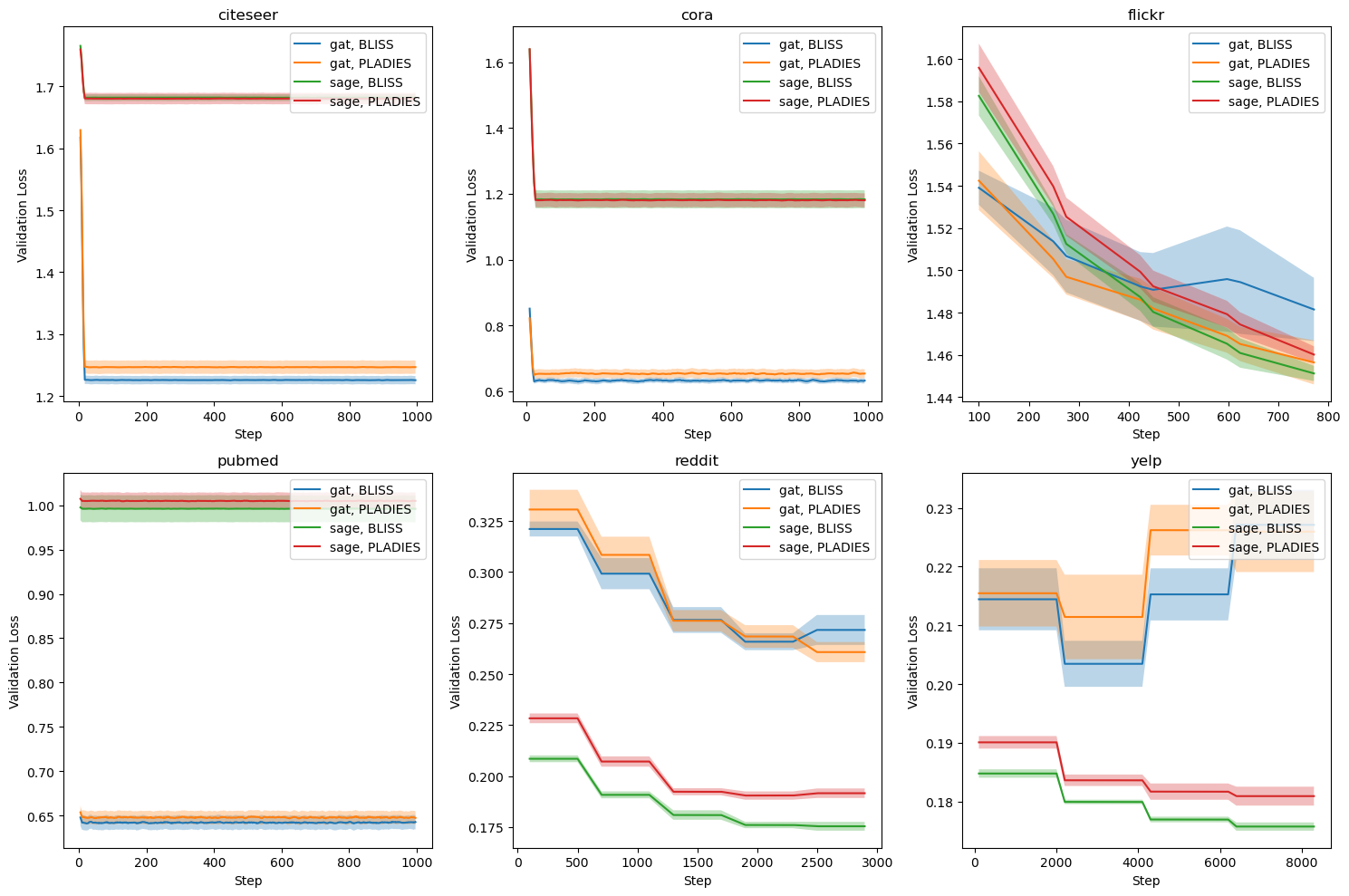}}
    \caption{Validation Loss for the same datasets and models as in \cref{fig:val_f1}. The figure illustrates the loss trends during training, averaged over 5 runs. The shaded regions represent the standard deviation across runs.}
    \label{fig:val_loss}
\end{figure}

\end{document}